\documentclass[journal]{IEEEtran}
\usepackage[latin1]{inputenc}
\usepackage[dvips]{graphicx}

\usepackage{algorithm}
\usepackage{algorithmic}
\usepackage{color}
 \usepackage{amsmath,amsfonts}
\newenvironment{definition}[1][Definition]{\begin{trivlist}
\item[\hskip \labelsep {\bfseries #1}]}{\end{trivlist}}

\begin{document}

\title{Conceptualization of seeded region growing by pixels aggregation. Part 2: how to localize a final partition invariant about the seeded region initialisation order.}
\author{Vincent Tariel }
\maketitle
\begin{abstract}
 In the previous paper, we have conceptualized the localization and the organization of seeded region growing by pixels aggregation (SRGPA) but we do not give the issue when there is a collision between two distinct regions during the growing process. In this paper, we propose two implementations to manage two classical growing processes: one without a boundary region region to divide the other regions and another with. Unfortunately, as noticed by Mehnert and Jakway (1997), this partition depends on the seeded region initialisation order (SRIO). We propose a growing process, invariant about SRIO such as the boundary region is the set of ambiguous pixels.
\end{abstract}

\begin{keywords}
Boundary, mathematical morphology, Minkowski addition, seeded region growing by pixel aggregation.
\end{keywords}
\IEEEpeerreviewmaketitle

\section{Introduction}
In the previous paper\cite{Tariel2008b}, we have conceptualized the localization and the organization of seeded region growing by pixels aggregation (SRGPA). This conceptualization has permitted to create a library dedicated to the implementation of algorithm using SRGPA (see annexe~\ref{ap:sum}). Each implementation using this library is quick and provides efficient algorithms.\\
At the end of most of algorithms using SRGPA, the regions are a partition of the space. In a classical growing process, there are two possible approaches to partition the space: one without a boundary region to divide the other regions, another with. Thanks to the conceptualization of SRGPA, it is easy to implement an algorithm such as the regions respect one or the other partitions at the end of the growing process. However, this partition depends on the seeded region initialisation order (SRIO)\cite{Beucher2004,Mehnert1997}. The localization of the inner border of each region depends on the SRIO. To overcome this problem, we define a set of ambiguous points. This set is called ambiguous points because in discrete space, there are some points such as it is impossible to determine to which regions they belong. We define a growing process that affect:
\begin{itemize}
\item the no ambiguous points to the appropriate regions,
\item the ambiguous points to the boundary region.
\end{itemize}
In this article, the notations are:
\begin{itemize}
\item let $E$ be a discrete space\footnote{ The space $E$, is a n-dimensional discrete space $\mathbb{Z}^n$, consisting of lattice points which coordinates are all integers in a three-dimensional Euclidean space $\mathbb{R}^n$. The elements of a n-dimensional image array are called points.},
\item let $\Omega$ be a domain of $E$  and $I$ its characteristic function such as $\Omega=\{\forall x\in E: I(x)\neq 0\}$,
\end{itemize}
Using this growing process, the localization of final partition is invariant about the SRIO.\\
The outline of the rest of the paper is as follows: in Sec.~II, we present the two classical growing processes.  In Sec.~III, we explain how to implement a growing process invariant about the SRIO. In Sec.~V, we make concluding remarks. 

\section{Classical growing processes}
This section presents two classical growing processes. For the first, there is no boundary region to divide the other regions. For the second, there is a boundary region to divide the other regions. The geodesic dilatation\cite{Schmitt1989} is used like an example but this approach can be used for the most of algorithms using SRGPA if the algorithm can be reduced in a succession of  geodesic dilatations\cite{Najman1996}.  This section is decomposed in two parts: definition of two distinct partitions and how to get both partitions for algorithms using SRGPA.
\subsection{Two distinct partitions}A segmentation of $\Omega$ is a simple-partition of $\Omega$ into subsets $X_i$ , $i= 1,\ldots,m$, for some $m$ if:
\begin{enumerate}
\item $\Omega=\cup_{i=1}^m X_i$
\item $\forall i\neq j \Rightarrow X_i\cap X_j = \emptyset$
\end{enumerate}
A segmentation of $\Omega$ is a $V$-boundary-partition\footnote{A $V$-boundary-partition is also a simple-partition.} of $\Omega$ into subsets $X_i$  $i= 1,\ldots,m$, for some $m$, and $X_b$  if:
\begin{enumerate}
\item $\Omega=(\cup_{i=1}^m X_i)\cup X_b$
\item $\forall i\neq j \Rightarrow (X_i\oplus V)\cap X_j = \emptyset$
\item $X_b\ominus V=\emptyset$
\end{enumerate}
The second condition defines that the boundary region divides the other regions and the third condition defines that the boundary region thickness is equal to 1.
\subsection{Simple-partition}
\begin{figure}
\begin{center}
\includegraphics[width=3cm]{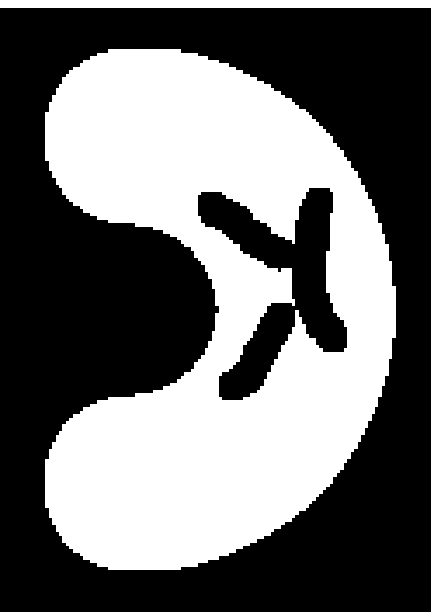}\includegraphics[width=3cm]{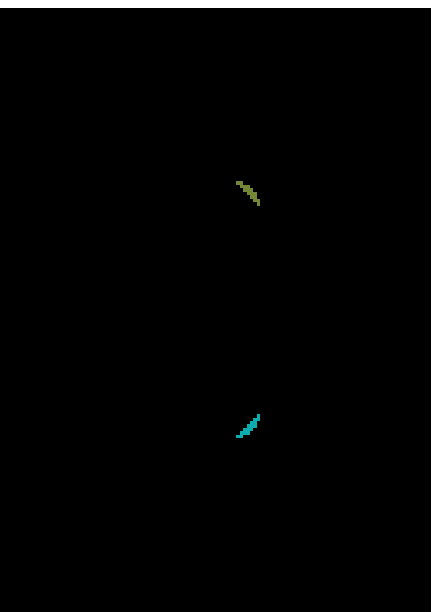}\includegraphics[width=3cm]{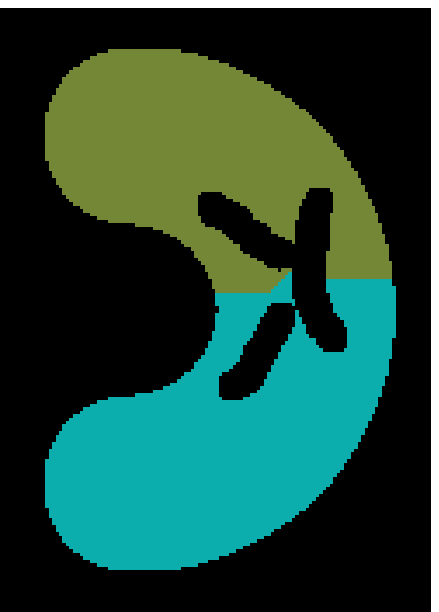}
\includegraphics[width=3cm]{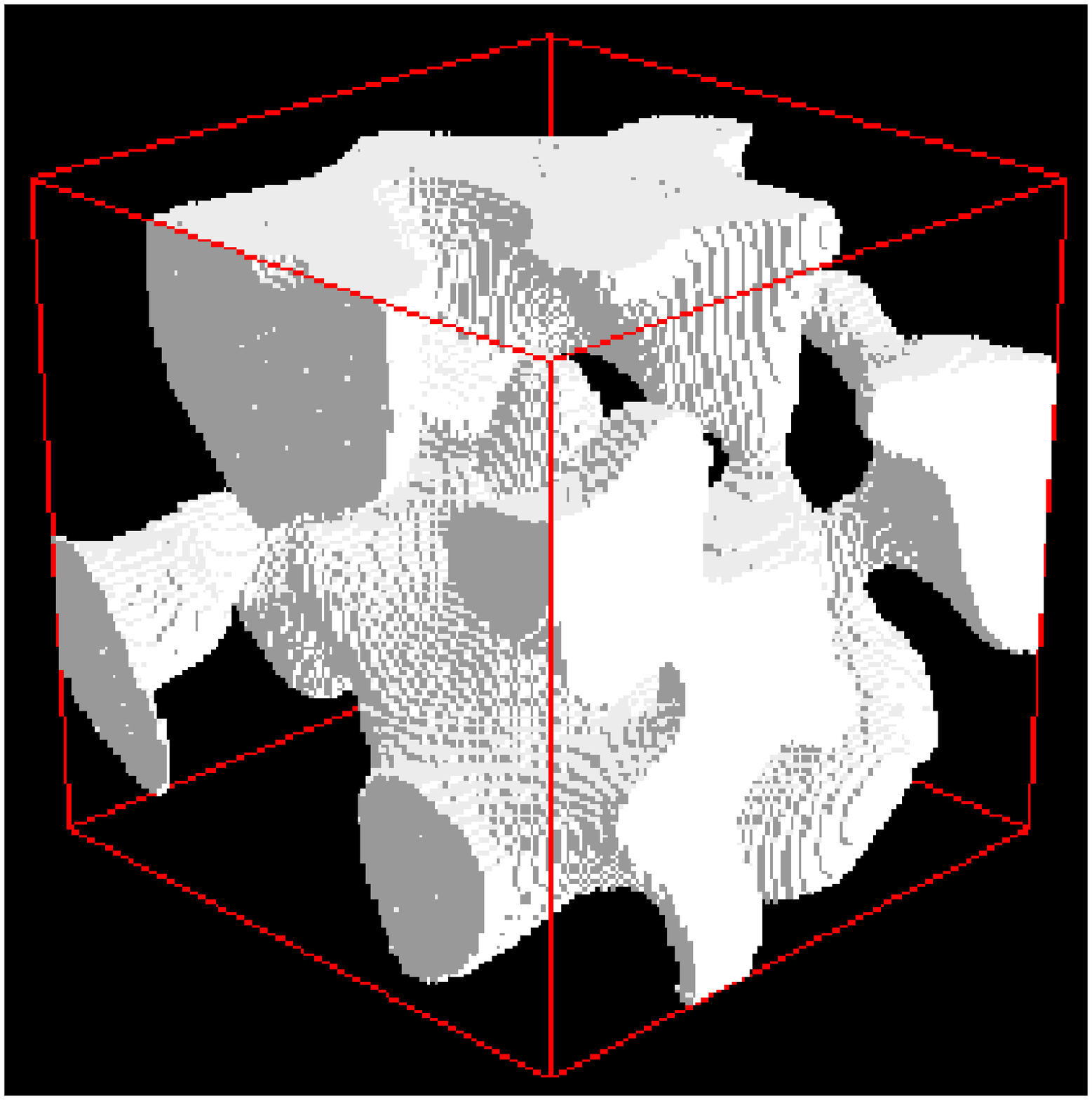}\includegraphics[width=3cm]{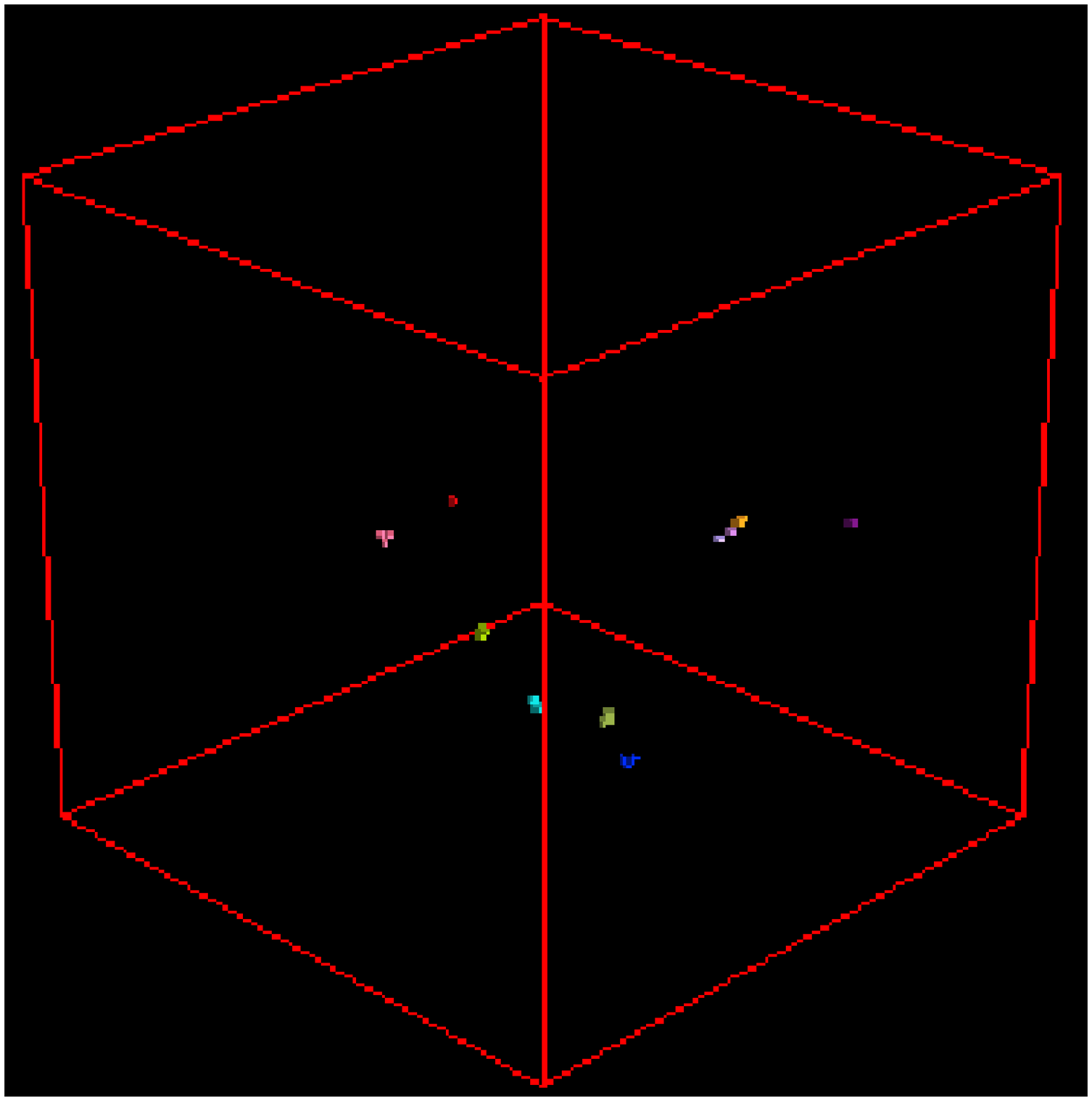}\includegraphics[width=3cm]{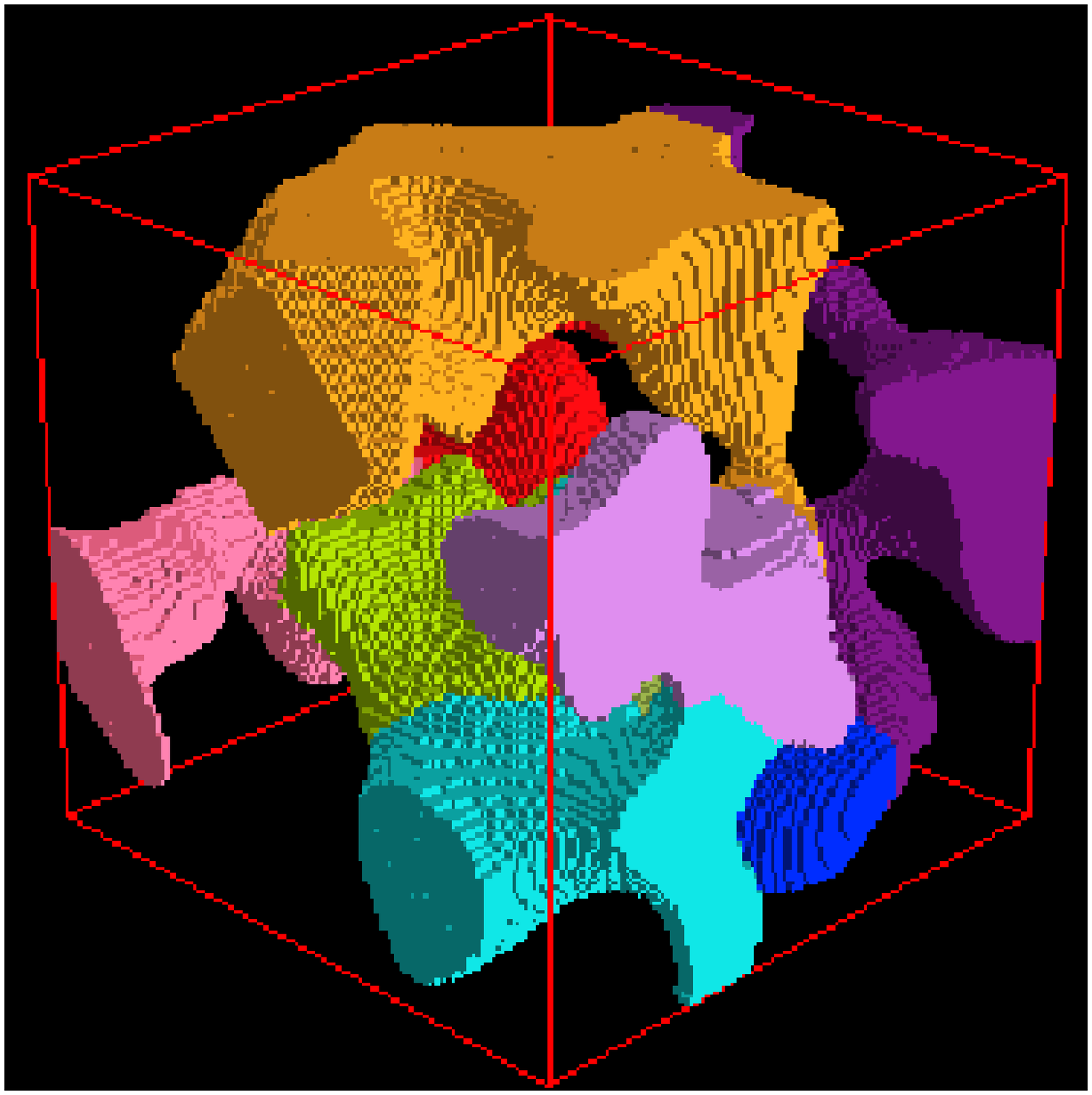}\\
\caption{For both series, the first image is the initial image, the second image is the seeds and the last image is the simple-partition after the geodesic dilatation. The first serie is the case in 2D and the second in 3D. For  both, the regions are a simple-partition of $\Omega=\{\forall x\in E: I(x)\neq 0\}$}
\label{vicord}
\end{center} 
\end{figure}

To get a simple-partition using the SRGPA, the zone of influence (ZI) at each region is localized on the outer boundary region excluding all other regions: $Z^t_i=(X_{i}^t \oplus V)\setminus (\bigcup\limits_{j \in \mathbb{N}} X_{j})$.  During the growing process, when a couple $(x,i)$ is extracted from the SQ, there is a simple growth: p.growth(x, i)). At the end of the growing process, the regions $X_i^{t=\infty}$ $i= 1,\ldots,m$ are a simple-partition of $\Omega$. The algorithm~\ref{alg0} is an example (see figure~\ref{vicord}).
 \begin{algorithm}[h!tp]
\caption{Geodesic dilatation}
\label{alg0}
\algsetup{indent=1em}
\begin{algorithmic}[20]
 \REQUIRE $I$, $S$ , $V$ \textit{//The binary image, the seeds, the neighbourhood}
\STATE \textit{// initialization}
\STATE System$\_$Queue s$\_$q( $\delta(x,i)=0\mbox{ if }I(x)\neq 0, OUT\mbox{ else}$, FIFO, 1); \textit{//A single FIFO queue such as if $I(x)=0$ then $(x,i)$ is not pushed in the SQ.}
\STATE Population p (s$\_$q); \textit{//create the object Population}
\STATE \fbox{\textbf{Restricted $N=\mathbb{N}$; }}
\STATE Tribe active(V, N);
\FORALL{$\forall s_i\in S$} 
\STATE int ref$\_$tr   = p.growth$\_$tribe(actif); \textit{//create a region/ZI, $(X^t_i,Z^t_i)$ such as \fbox{$Z^t_i=(X_{i}^t \oplus V)\setminus (\bigcup\limits_{j \in \mathbb{N}} X_{j})$}}
\STATE  p.growth($s_i$, ref$\_$tr ); 
\ENDFOR
\STATE \textit{//the growing process}
\STATE  s$\_$q.select$\_$queue(0); \textit{//Select the single FIFO queue.}
\WHILE{s$\_$q.empty()==false}
\STATE  $(x,i)=s$\_$q$.pop();
\STATE \fbox{\textbf{ p.growth(x, i ); }}
\ENDWHILE
\RETURN p.X();
\end{algorithmic}
 \end{algorithm}

\subsection{The $V$-boundary-partition}
\begin{figure}
\begin{center}
\includegraphics[width=3cm]{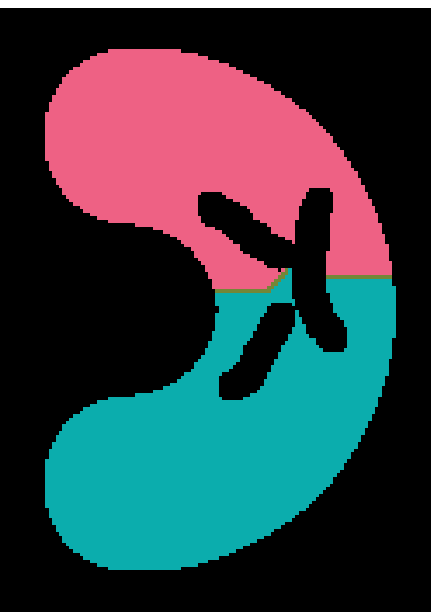}\includegraphics[width=3cm]{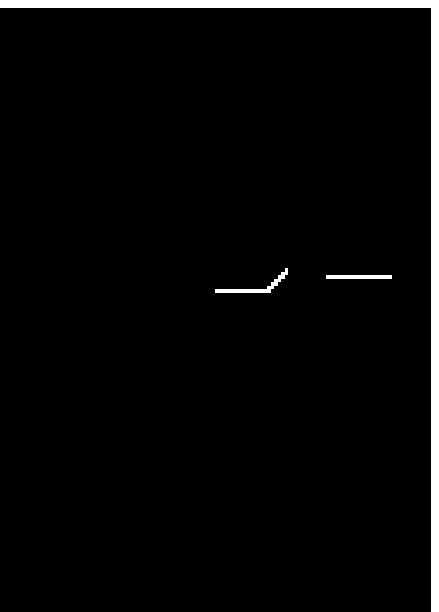}\includegraphics[width=3cm]{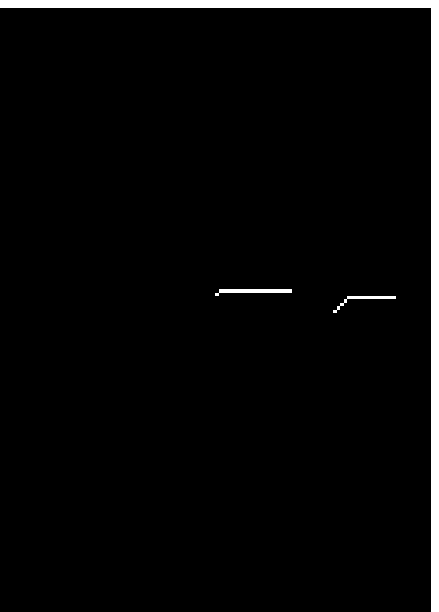}\\
\includegraphics[width=3cm]{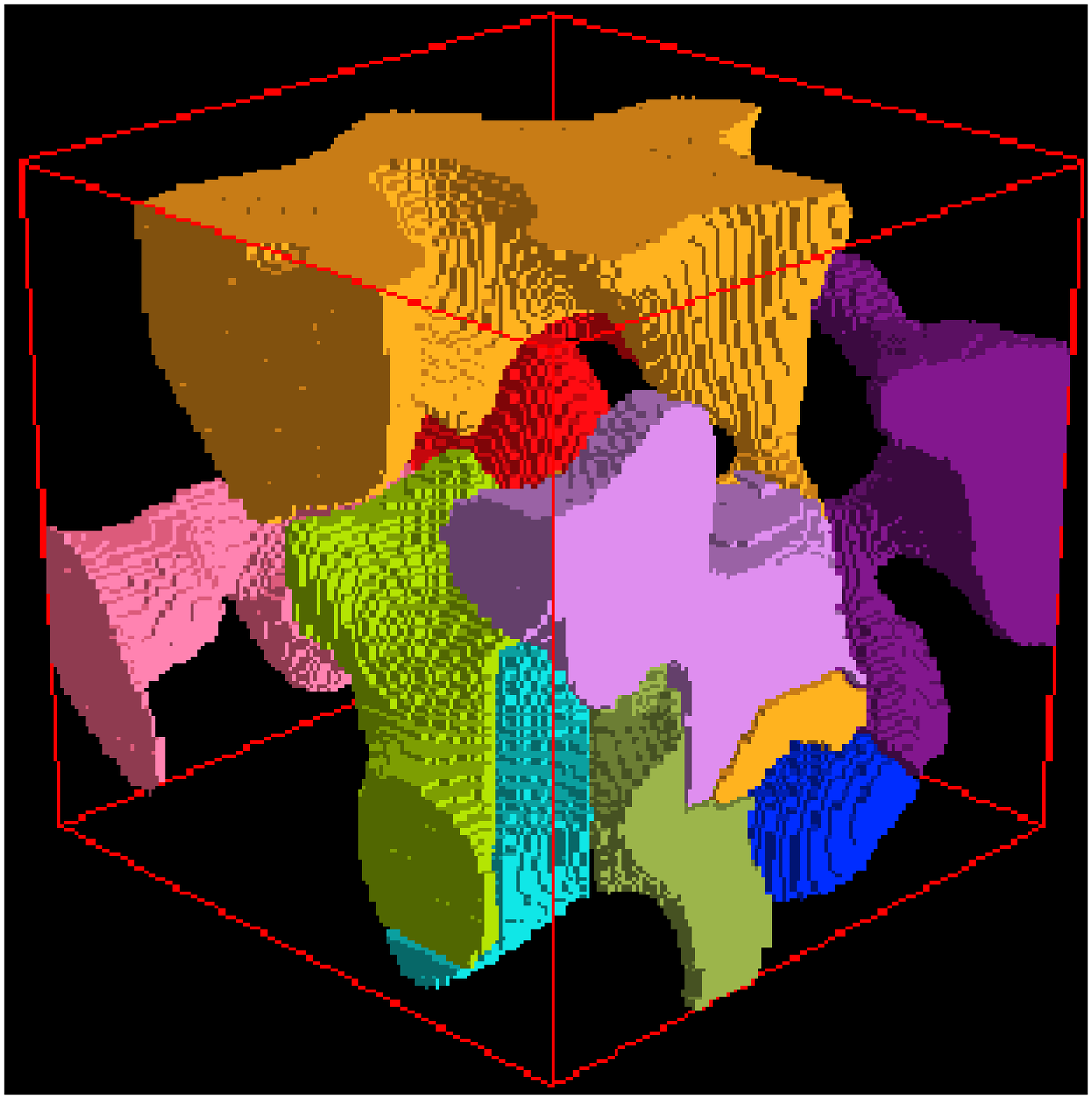}\includegraphics[width=3cm]{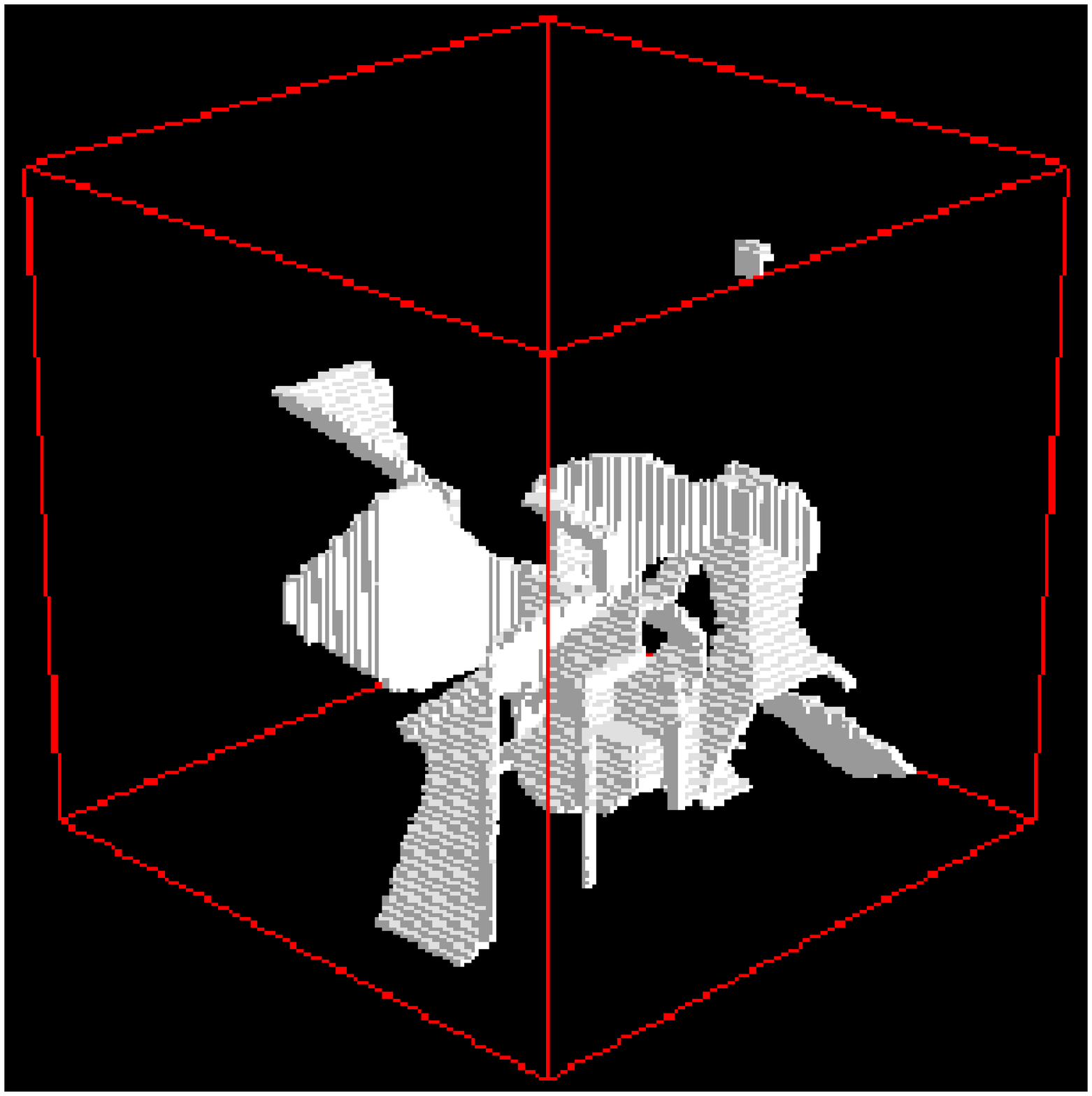}\includegraphics[width=3cm]{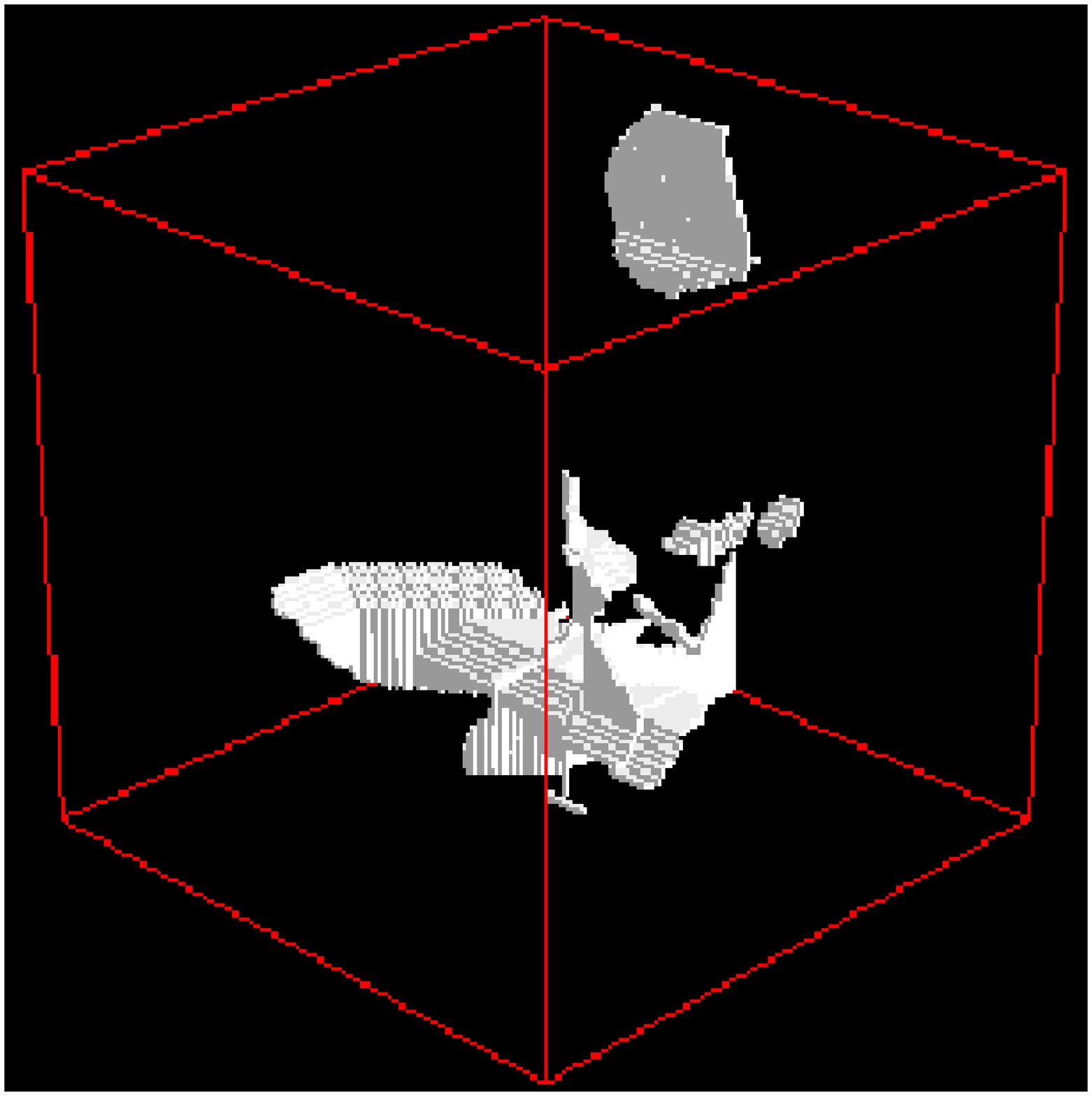}
\caption{For both series, the first image is $V$-boundary-partition obtained by the geodesic dilatation with an boundary, the second figure and third fiigure are the visualisation of the boundary region depending on the choosen neighbourhood. For the second figure, it is the 8-neightborhood in 2D and 26-neightborhood in 3D and for the third figure, it is  4-neightborhood in 2D and 6-neightborhood in 3D.}
\label{vicord_b}
\end{center} 
\end{figure}

To get a simple-partition using the SRGPA, a boundary region, $X_b$, is added such as its ZI is always empty. For all the regions except the boundary region,  their ZI are localized on the outer boundary region excluding  all the regions: $Z^t_i=(X_{i}^t \oplus V)\setminus (\bigcup\limits_{j \in \mathbb{N}} X_{j})$. The simple growth, p.growth( x, i ), is substituted by 
\begin{itemize}
\item  if there is more than 2 ZI  on x, then growth on $x$ of the boundary region,
\item else the growth on $x$ of the region $i$
\end{itemize}
Using this definition, at the end of the growing process, the regions $X_i^{t=\infty}$ $i= 1,\ldots,m$, and $X_b$ are a $V$-boundary-partition of $\Omega$. The algorithm~\ref{alg1} is an example (see figure~\ref{vicord_b}).
 \begin{algorithm}[h!tp]
\caption{Geodesic dilatation with an boundary}
\label{alg1}
\algsetup{indent=1em}
\begin{algorithmic}[20]
 \REQUIRE $I$, $S$ , $V$ \textit{//The binary image, the seeds, the neighbourhood}
\STATE \textit{// initialization}
\STATE System$\_$Queue s$\_$q( $\delta(x,i)=0\mbox{ if }I(x)\neq 0, OUT\mbox{ else}$, FIFO, 1); \textit{//A single FIFO queue such as if $I(x)=0$ then $(x,i)$ is not pushed in the SQ.}
\STATE Population p (s$\_$q); \textit{//create the object Population}
\STATE \fbox{\textbf{Tribe passive($V=\emptyset$);}}
\STATE \textit{//create a boundary region/ZI, $(X^t_b,Z^t_b)$ such as $Z^t_i=\emptyset$}
\STATE \fbox{\textbf{int ref$\_$boundary   = p.growth$\_$tribe(passive);}}
\STATE \fbox{\textbf{Restricted $N$=$\mathbb{N}$; }}
\STATE Tribe active(V, N);
\FORALL{$\forall s_i\in S$} 
\STATE int ref$\_$tr   = p.growth$\_$tribe(actif); \textit{//create a region/ZI, $(X^t_i,Z^t_i)$ such as \fbox{$Z^t_i=(X_{i}^t \oplus V)\setminus (\bigcup\limits_{j \in \mathbb{N}} X_{j})$}}
\STATE  p.growth($s_i$, ref$\_$tr ); 
\ENDFOR
\STATE \textit{//the growing process}
\STATE  s$\_$q.select$\_$queue(0); \textit{//Select the single FIFO queue.}
\WHILE{s$\_$q.empty()==false}
\STATE  $(x,i)=s$\_$q$.pop();
\IF{pop.Z()[x].size()$\geq$2}
\STATE \textit{//growth of the boundary region}
\STATE \fbox{\textbf{ p.growth(x, ref$\_$boundary); }}
\ELSE
\STATE \fbox{\textbf{ p.growth(x, i ); }}\textit{//simple growth}
\ENDIF
\ENDWHILE
\RETURN p.X();
\end{algorithmic}
 \end{algorithm}

\subsection{The partition depends on SRIO}
\begin{figure}
\begin{center}
\fbox{\includegraphics[height=3cm]{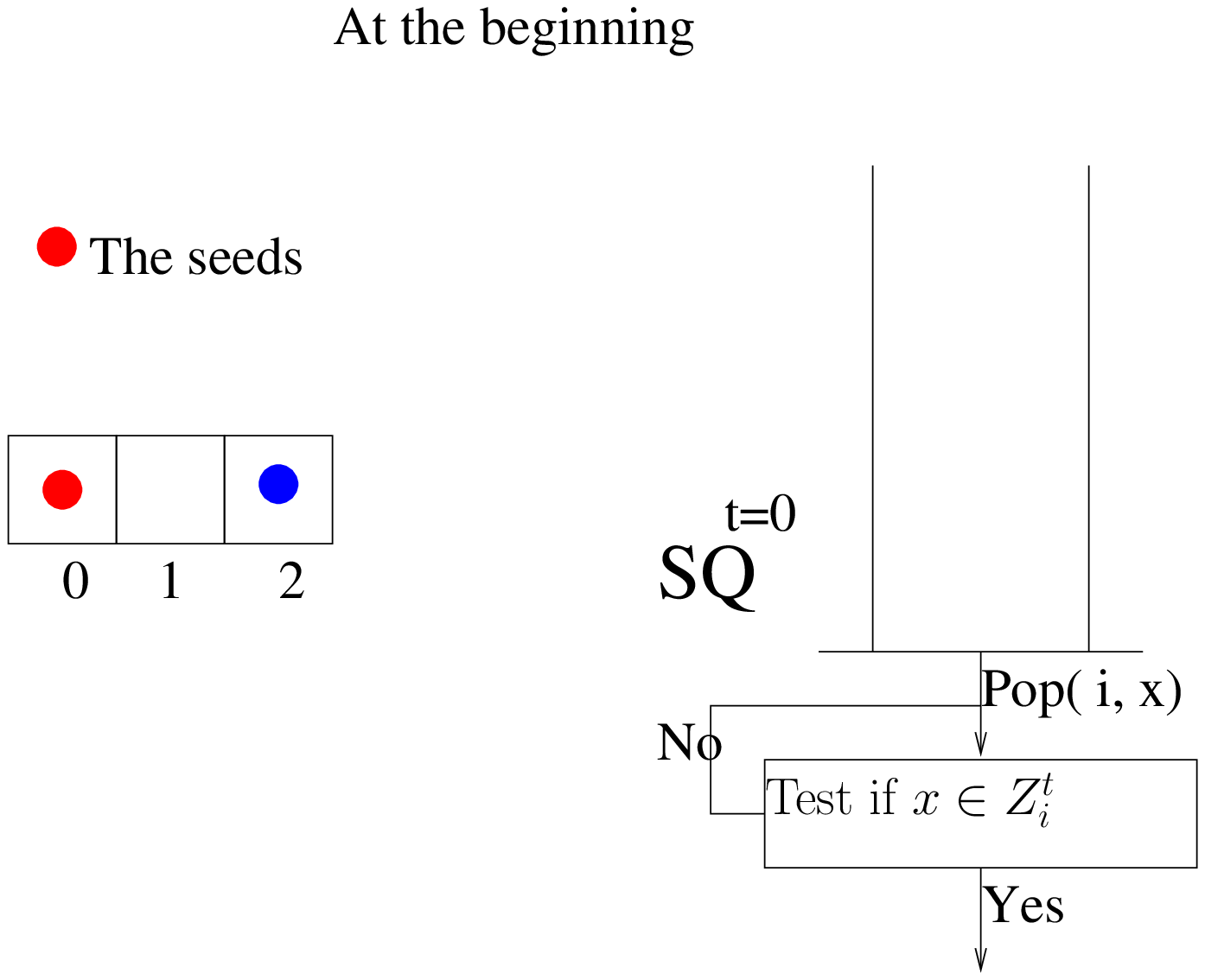}}
\fbox{\includegraphics[height=3cm]{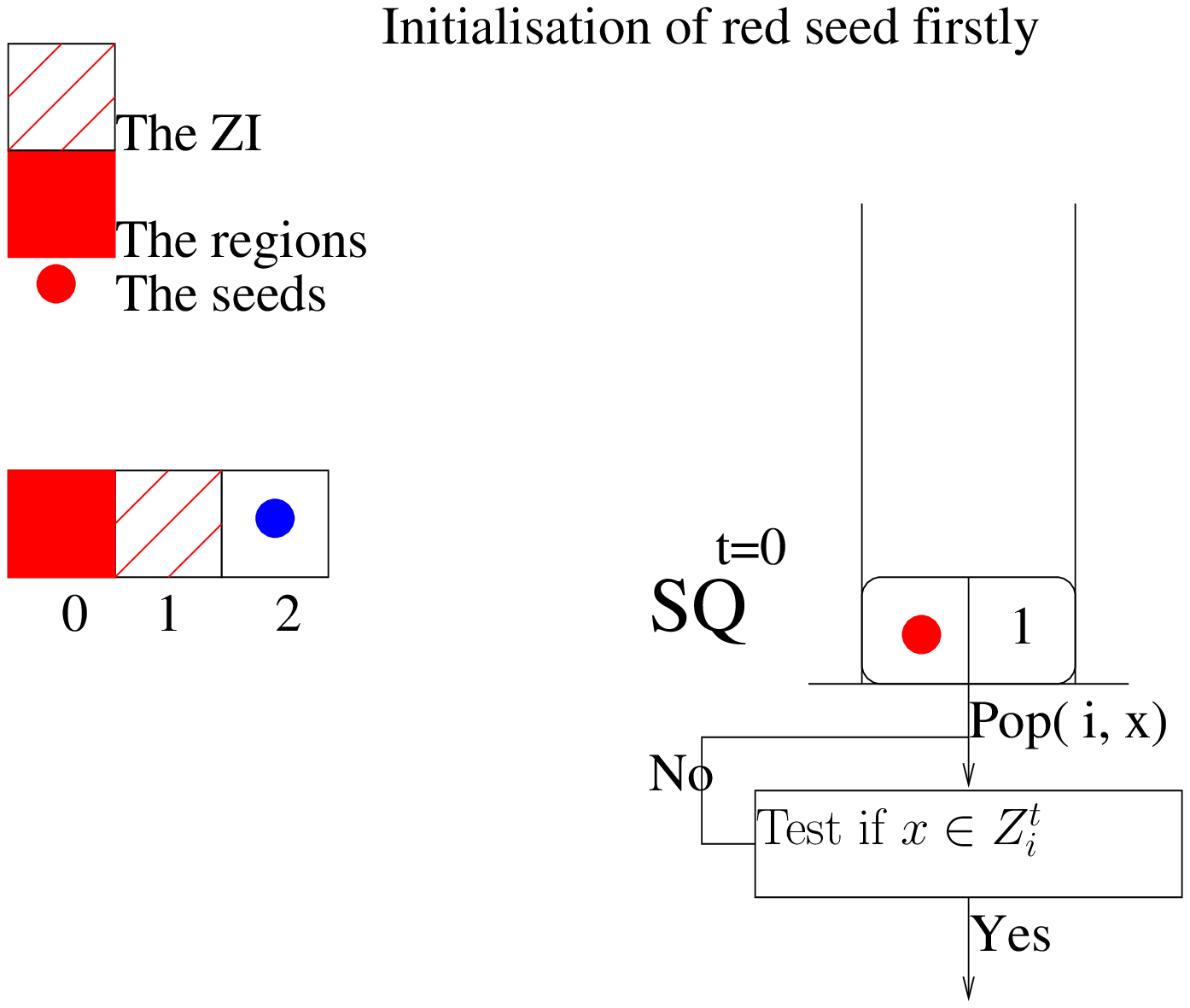}}
\fbox{\includegraphics[height=3cm]{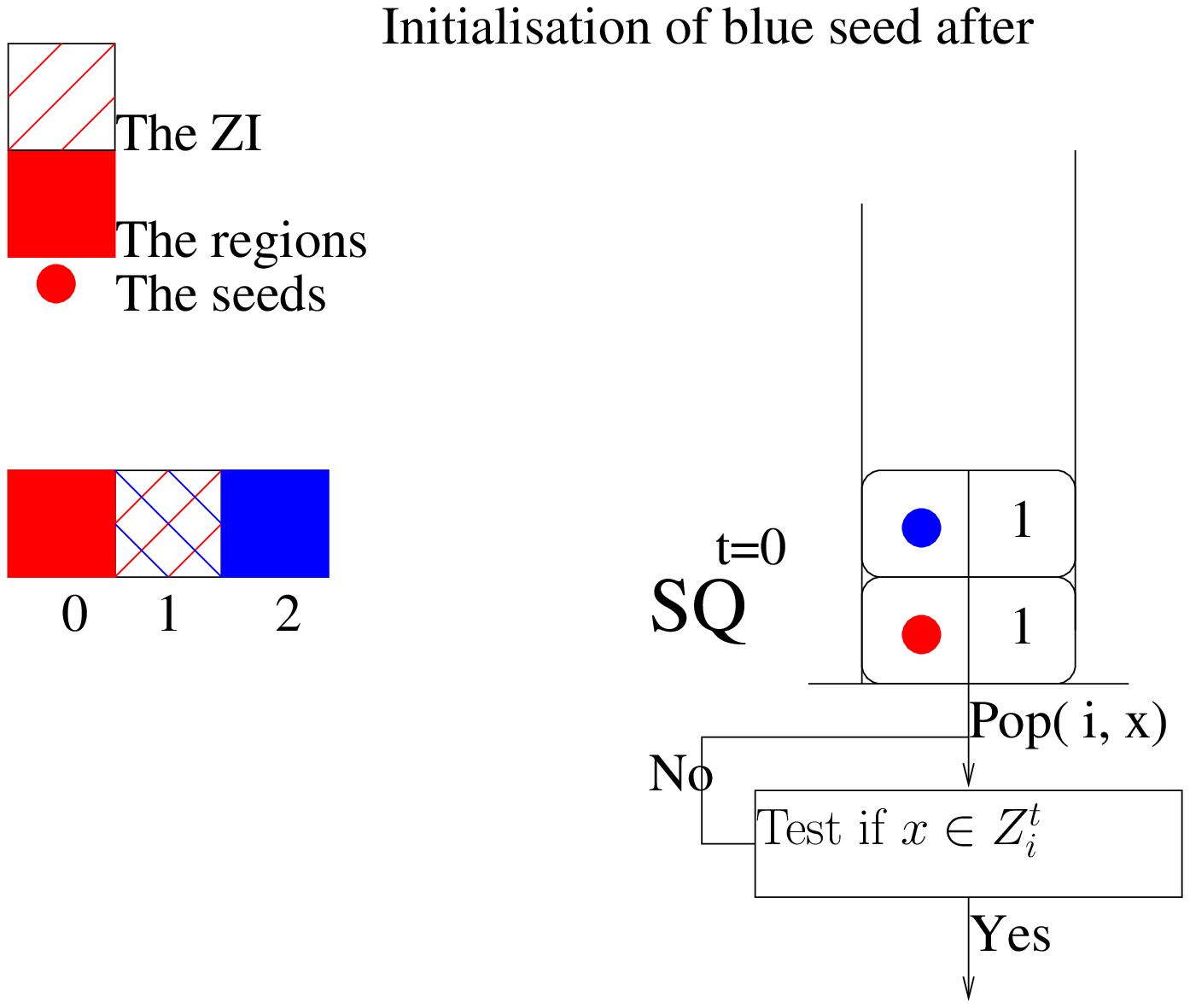}}
\fbox{\includegraphics[height=3cm]{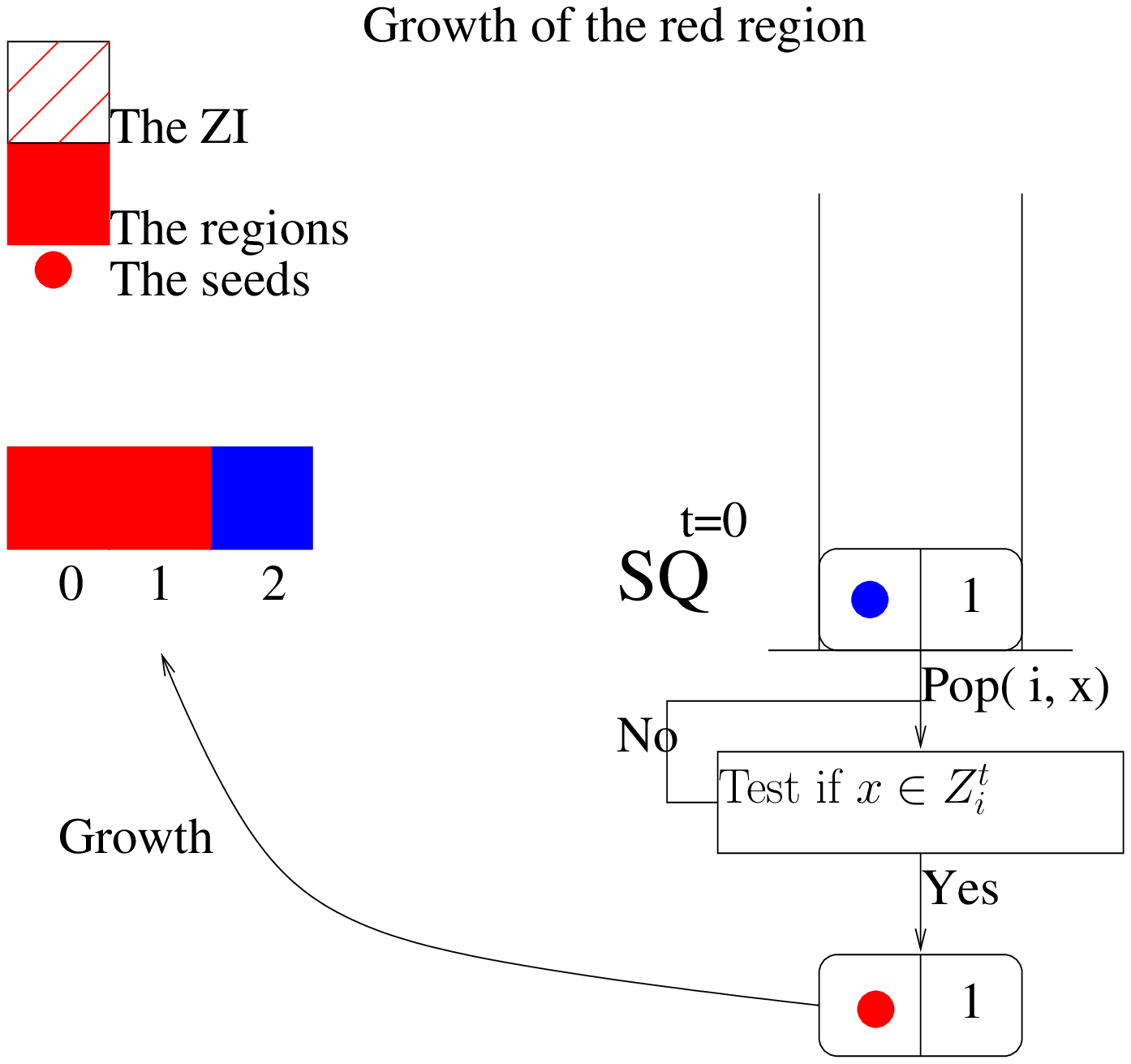}}
\fbox{\includegraphics[height=3cm]{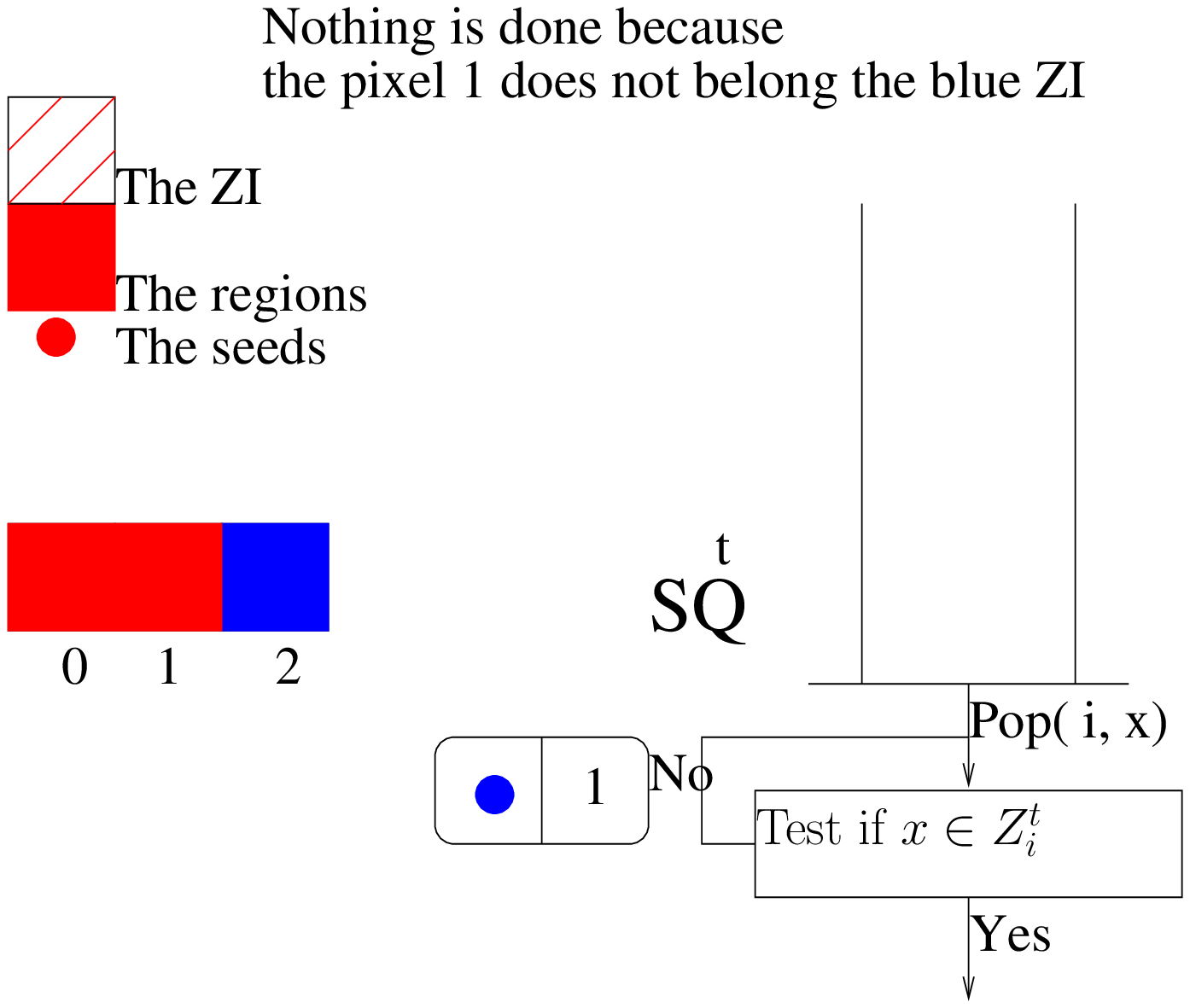}}
\caption{This serie shows the geodesic dilatation without a boundary region to divide the other regions such as the red seed is initialized firstly. The point 1 is an ambiguous point  in this growing process because  this point belongs to the region initialized firstly. In this case, it is the red region.}
\label{init}
\end{center} 
\end{figure}

\begin{figure}
\begin{center}
\fbox{\includegraphics[height=3cm]{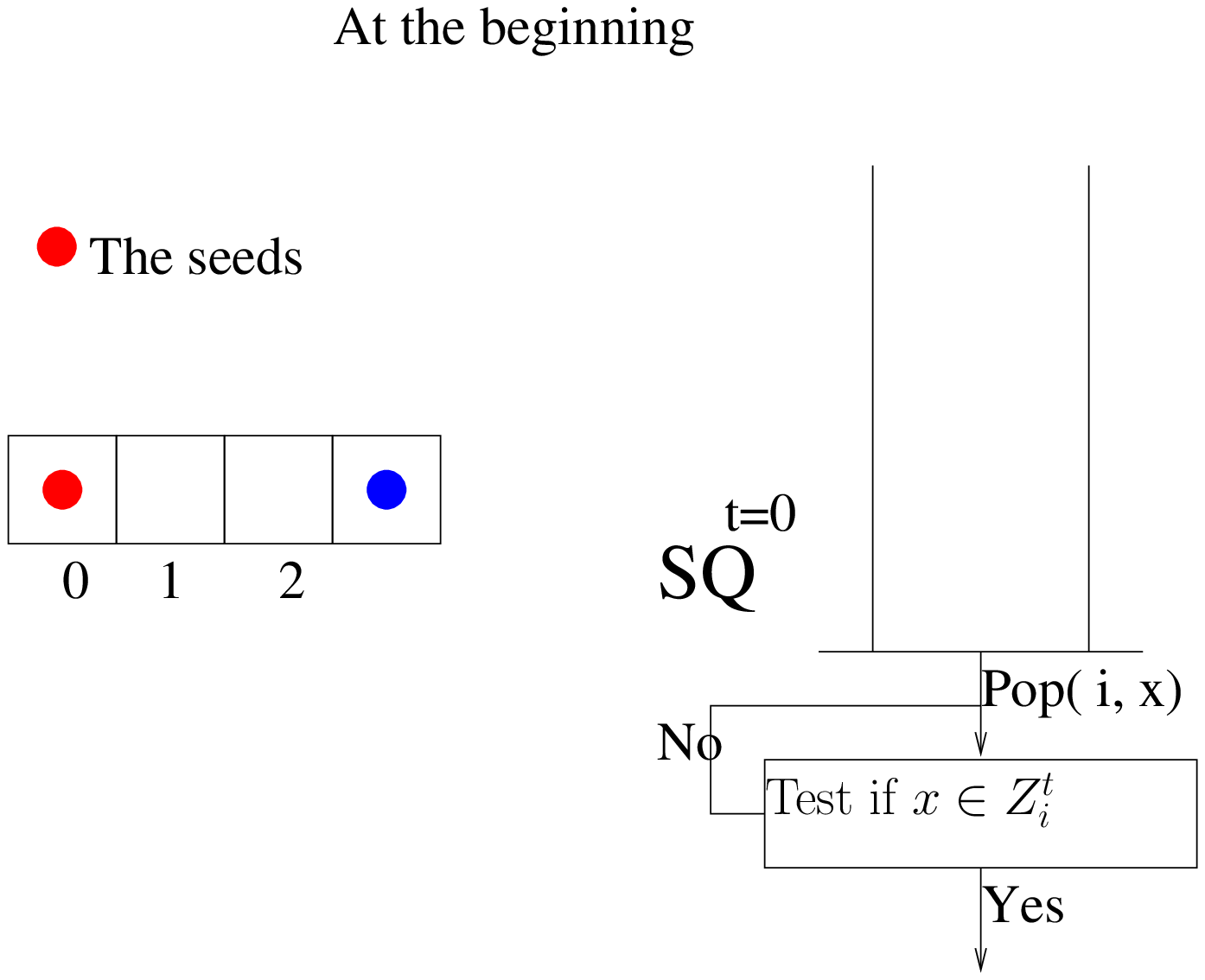}}
\fbox{\includegraphics[height=3cm]{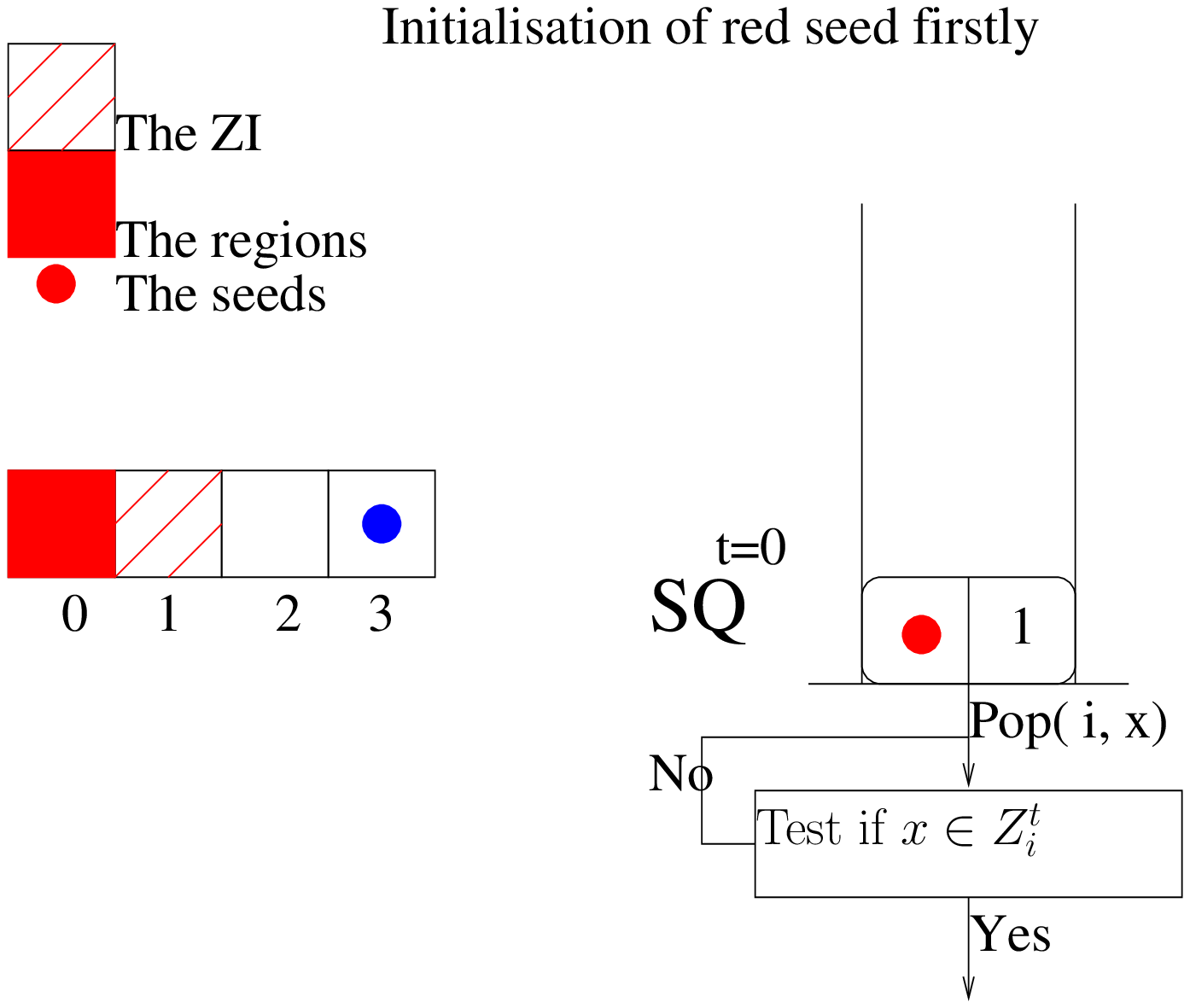}}
\fbox{\includegraphics[height=3cm]{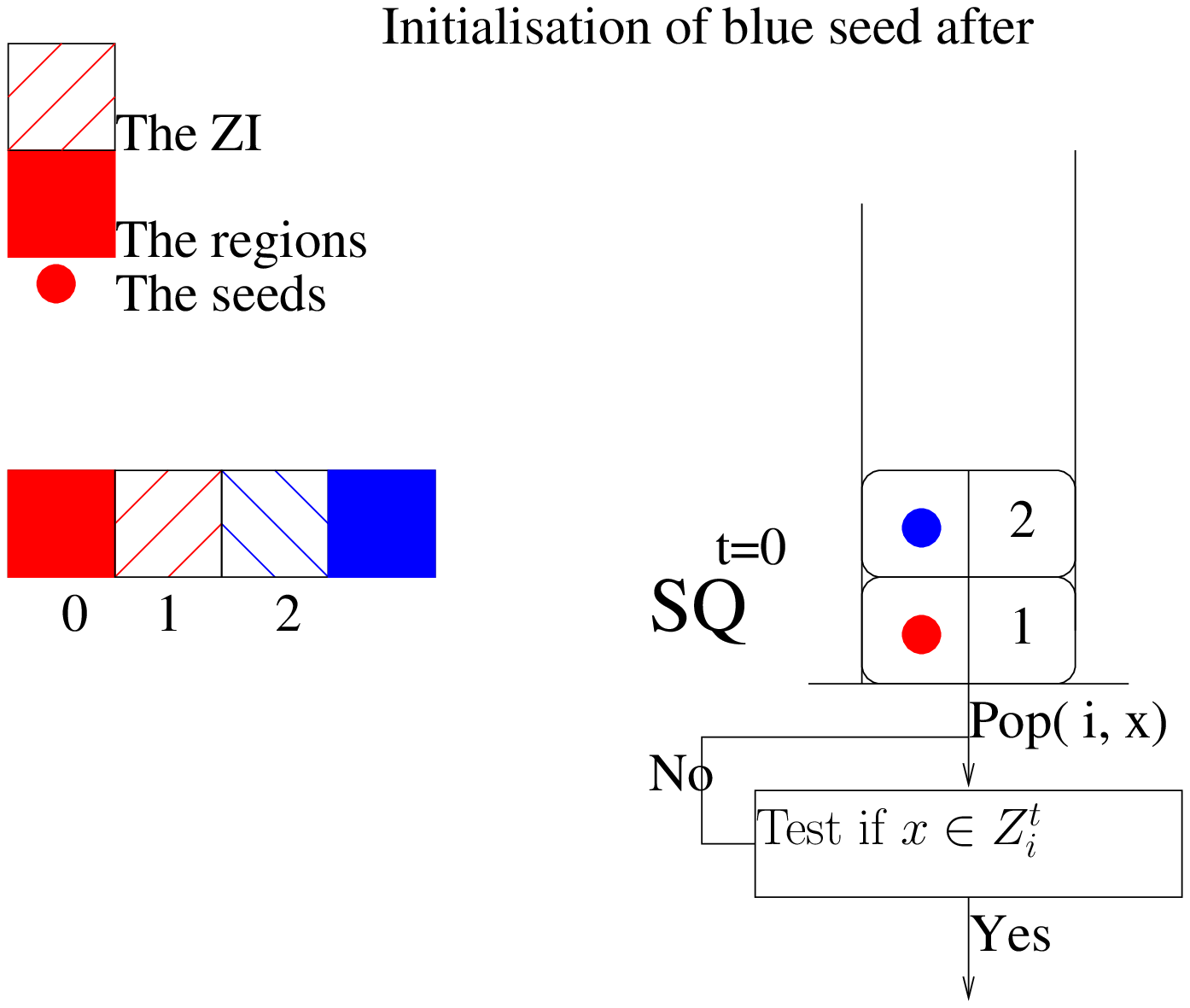}}
\fbox{\includegraphics[height=3cm]{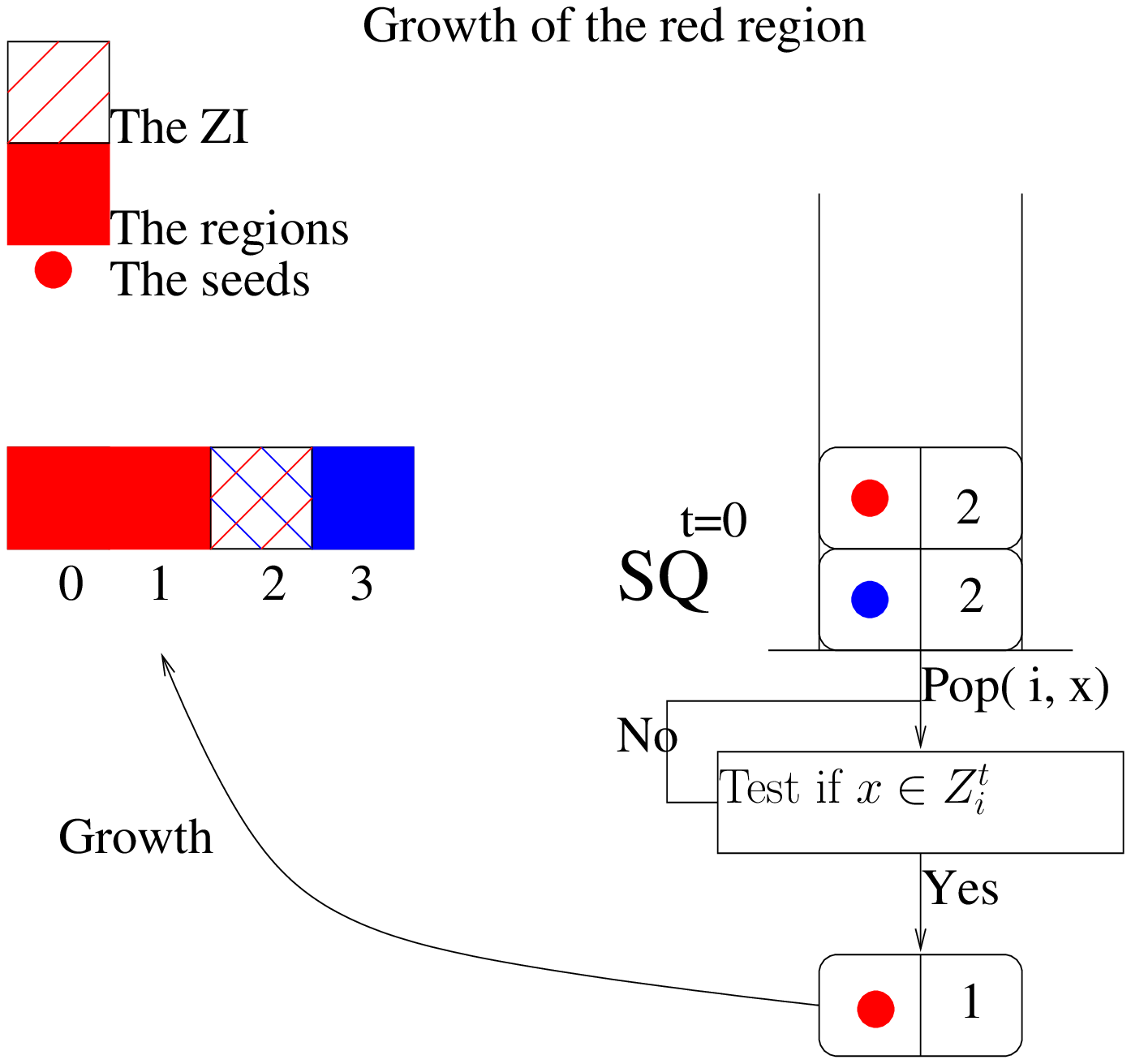}}
\fbox{\includegraphics[height=3cm]{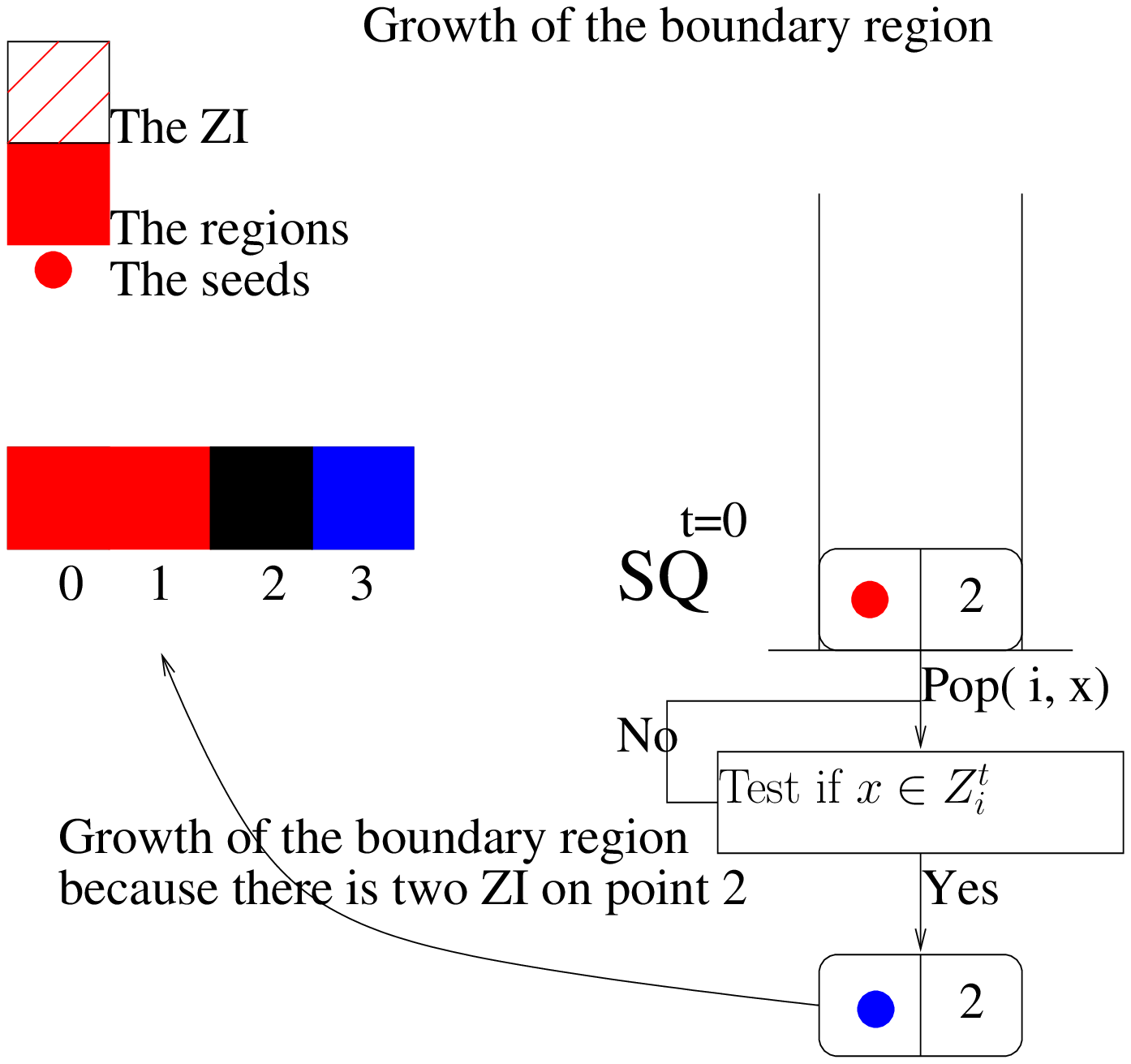}}
\fbox{\includegraphics[height=3cm]{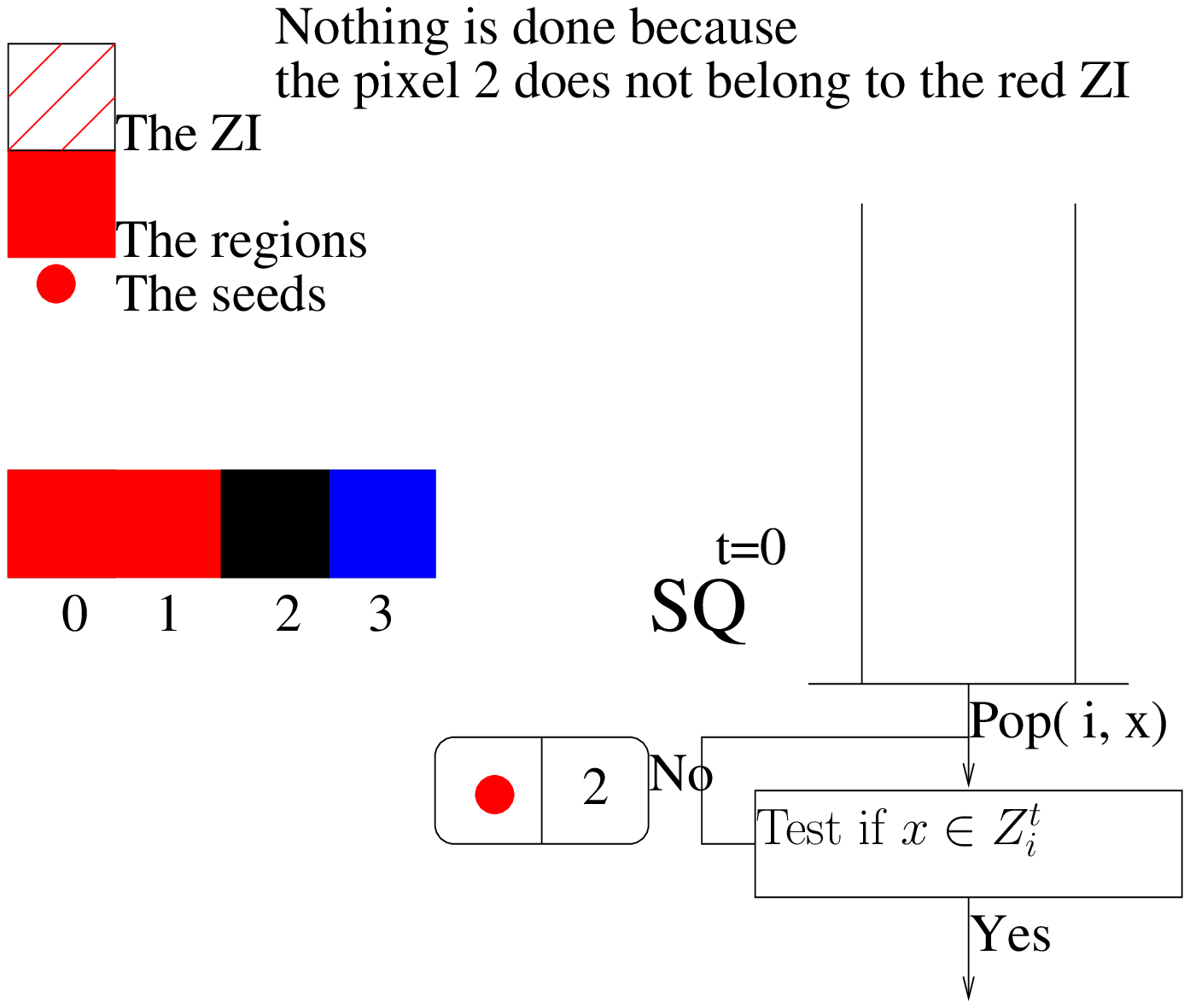}}
\caption{ This serie shows the geodesic dilatation with a boundary region to divide the other regions such as the red seed is initialized firstly. The point 1 and 2 are ambiguous pixels in this growing process because they belong to different region depending on SRIO. In this case, the point 1 belongs to the red region and the point 2 belongs to the boundary region but if the blue region is initialized at first, the point 1 will belong to the boundary region and the point 2 will belong to the blue region.}
\label{init2}
\end{center} 
\end{figure}

Whatever the growing process is, the final partition is not invariant about SRIO.  The figure~\ref{init} shows the case with an ambiguous pixel for the growing process without a boundary region to divide the other regions.  The figure~\ref{init2} shows the case with two ambiguous pixels for the growing process with a boundary  region to divide the other regions. 
The localization of the inner border of each region depends on SRIO. The next section proposes a solution to overcome this limitation.

\section{Invariance about the seeded initialisation order}
\subsection{Why is there dependence?}

\begin{definition}
Let $\Omega$ be a domain of $E$ and a and b two points of $\Omega$. We call
geodesic distance $d_\Omega(a, b)$ in $A$ the lower bound of the
length of the paths y in $\Omega$ linking a and b.\\
Let $s$  be a set. We call the geodesic distance $d_\Omega(s, b)=\min_{\forall a\in s}d_\Omega(a, b)$, the lower bound of all geodesic distance $d_\Omega(a, b)$ such as  $a$ belongs to $s$.\\
The geodesic influence zone\cite{Schmitt1989}, $z_A(s_i)$, of the seeds, $S=(s_i)_{1\leq i \leq n}$, of E in $\Omega$, is the set of the points of $\Omega$,
for which the geodesic distance to $s_i$ is smaller than the geodesic distance to other seeds of $S$.
\[
z_A(s_i)=\{\forall x\in \Omega:(\forall j\neq i \Rightarrow d_{\Omega}(s_i)< d_{\Omega}(s_j)\}
\] 
\end{definition}
The $z_A(s_i)_{1\leq i \leq n}$ is not a partition of $\Omega$ because $\cup_{i=1}^n z_A(s_i)\neq\Omega$.  In fact, it is possible to demonstrate that $\cup_{i=1}^n z_\Omega(s_i)=\Omega\uplus A$. The symbol $\uplus$ means the disjoint union:$ B\uplus C=\{B\cup C: B\cap C=\emptyset\}$. The set $A$, called ambigous points, is 
\begin{eqnarray*}
A=\{\forall x\in \Omega:\exists i,j \Rightarrow \\
( d_{\Omega}(s_i)= d_{\Omega}(s_j))\mbox{ and }\\
\left(  (\forall k \neq (\mbox{i and j})): d_{\Omega}(s_i)\leq d_{\Omega}(s_k) \right)  \}
\end{eqnarray*}
The set$A$ is  all the points of $\Omega$ for which the geodesic distance to $s_i$ and $s_{j\neq i}$ is equal and smaller than the geodesic distance to other seeds of $S$. The $z_A(s_i)_{1\leq i \leq n}$ and $A$ is a simple-partition of $\Omega$.  In the previous implementation of the geodesic dilatation, the ambiguous points are distributed depending on the seeded initialisation order (see figure~\ref{init} and~\ref{init2}). The next paragraph presents an implementation such as the boundary region is the set of ambiguous points. 

\subsection{Boundary as ambiguous points}
\begin{figure}
\begin{center}
\fbox{\includegraphics[width=1.4cm]{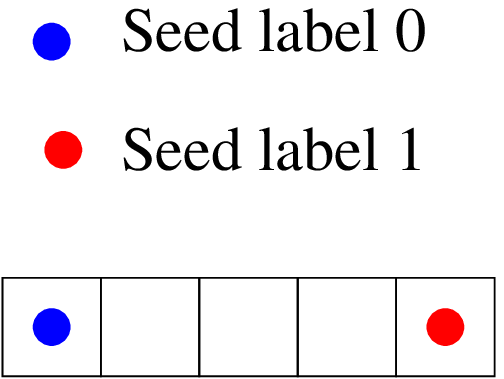}}
\fbox{\includegraphics[width=1.4cm]{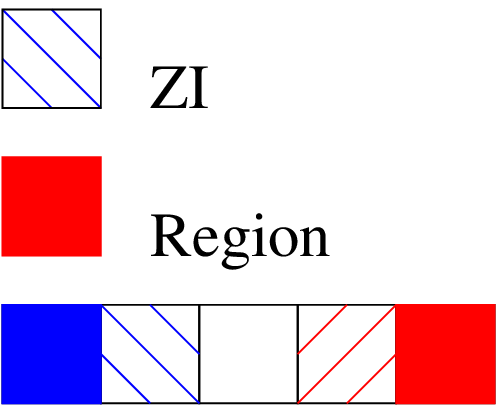}}
\fbox{\includegraphics[width=1.4cm]{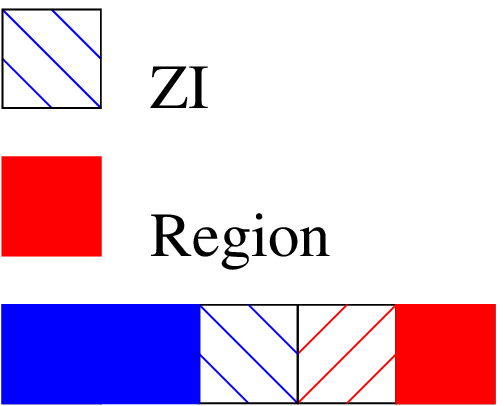}}
\fbox{\includegraphics[width=1.4cm]{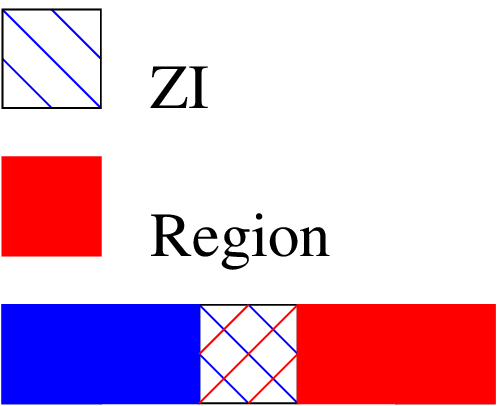}}
\fbox{\includegraphics[width=1.4cm]{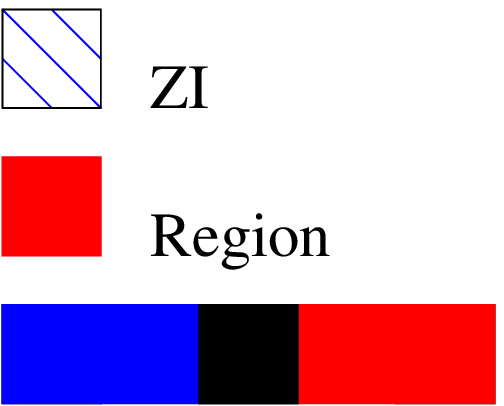}}
\fbox{\includegraphics[width=1.4cm]{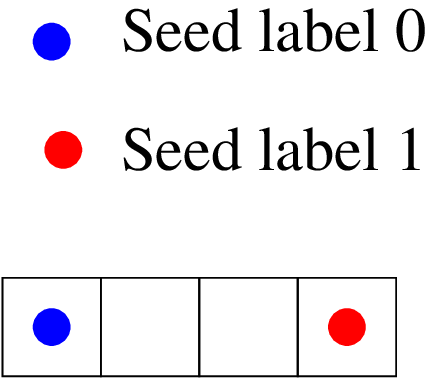}}
\fbox{\includegraphics[width=1.4cm]{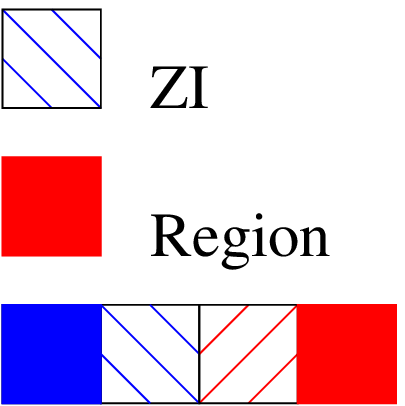}}
\fbox{\includegraphics[width=1.4cm]{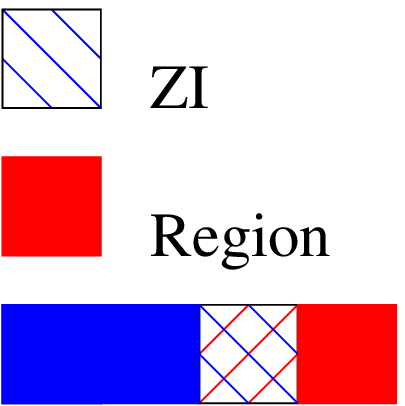}}
\fbox{\includegraphics[width=1.4cm]{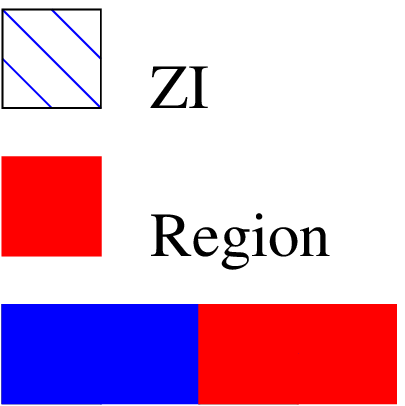}}
\caption{The first serie is the case of one ambiguous point. There is a classical growth until there are two ZI in the same point, $x$. In this case, the min$\_$elements( pop.Z()[x]) returns 0 because there are two ZI of label 0 and 1. There is the boundary growth because the couple extracted from the queue has a label 0 equal to min$\_$elements( pop.Z()[x]). The second serie is the case without ambiguous point. There is a classical growth until there is two ZI in the same point, $x$. The min$\_$elements( pop.Z()[x]) returns 0 because there are two ZI of label 0 and 1. There is the region growth of label 1 because the couple extracted from the queue has a label 1 not equal to min$\_$elements( pop.Z()[x]).}
\label{sqb}
\end{center} 
\end{figure}

\begin{figure}
\begin{center}
\includegraphics[width=3cm]{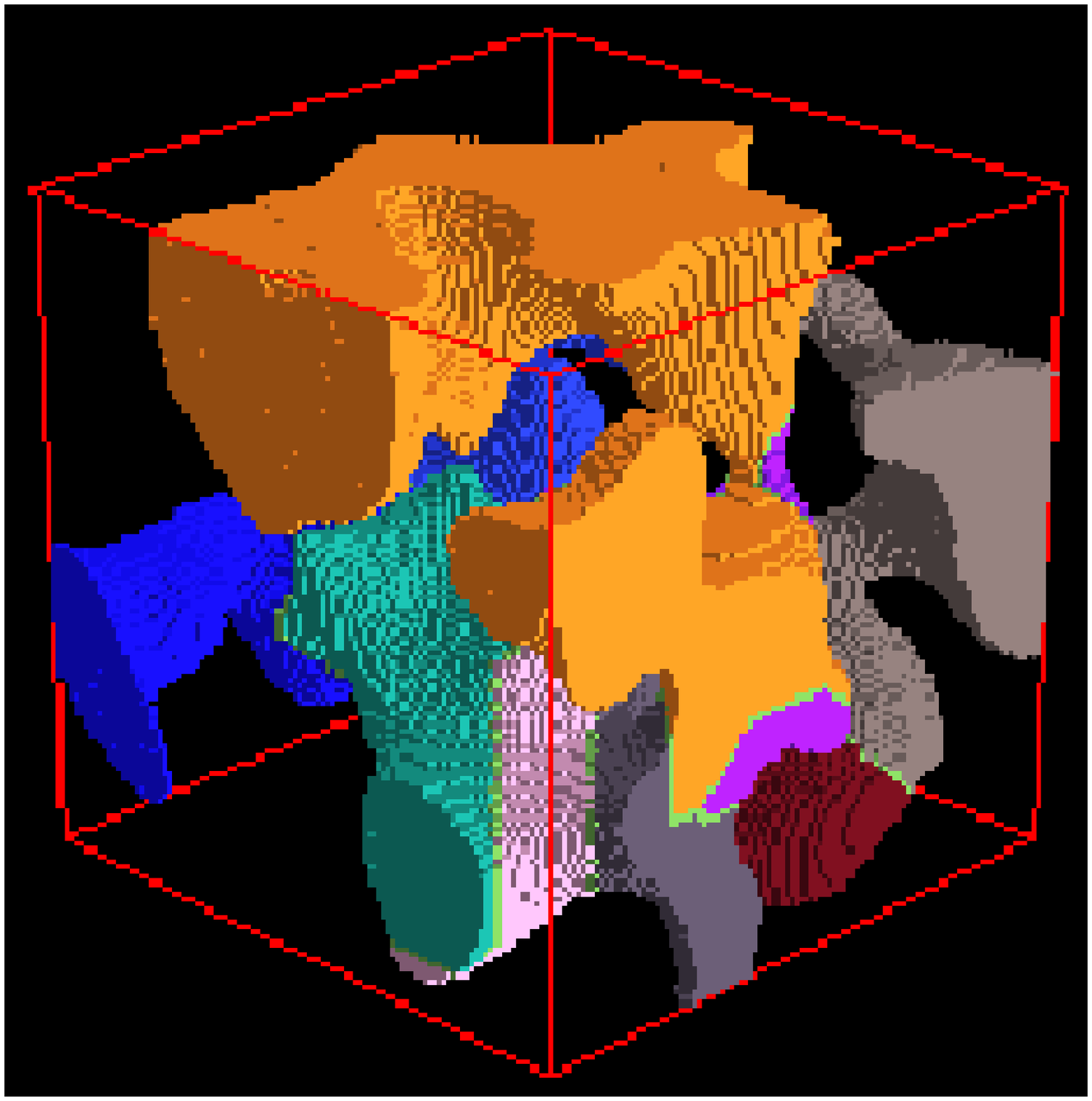}\includegraphics[width=3cm]{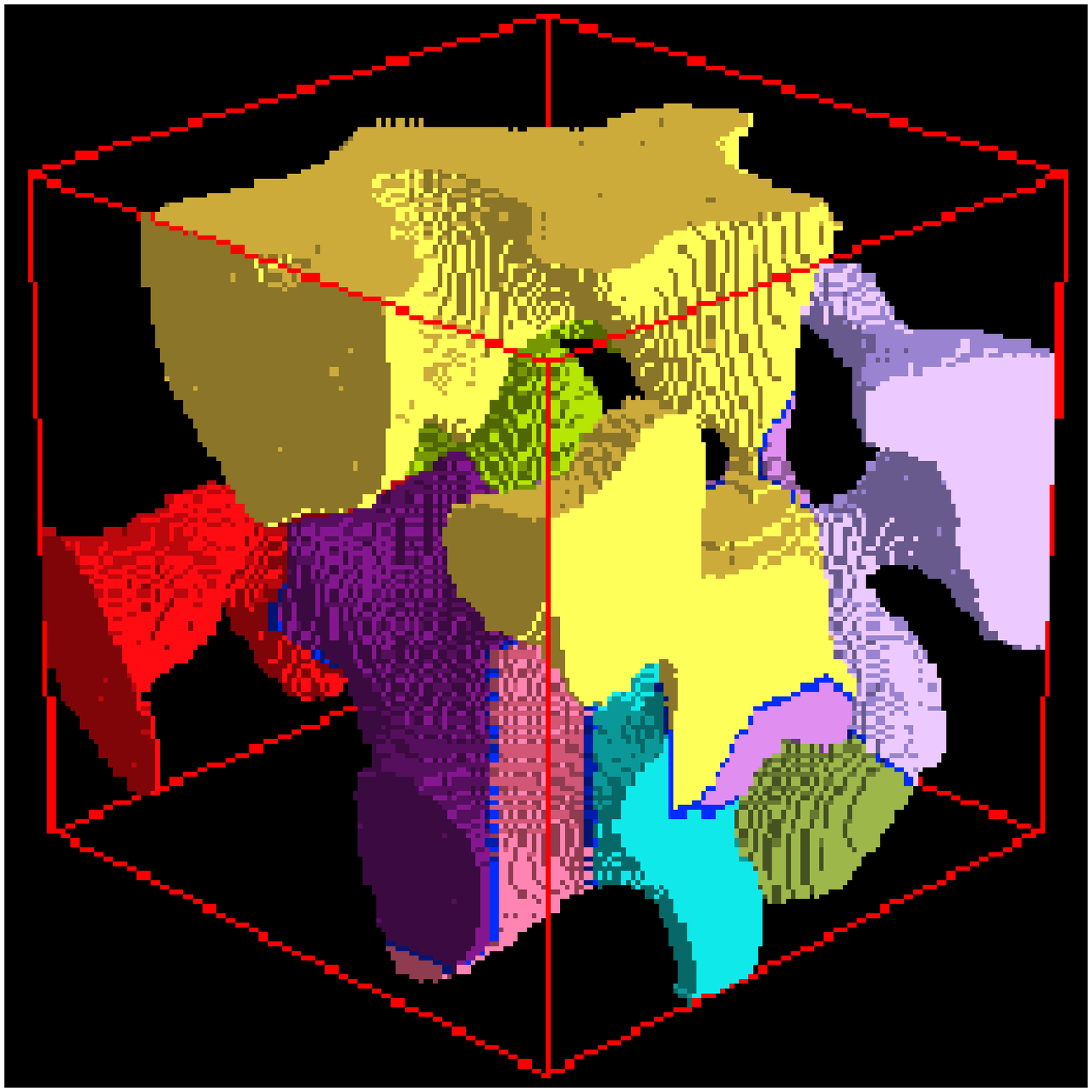}\includegraphics[width=3cm]{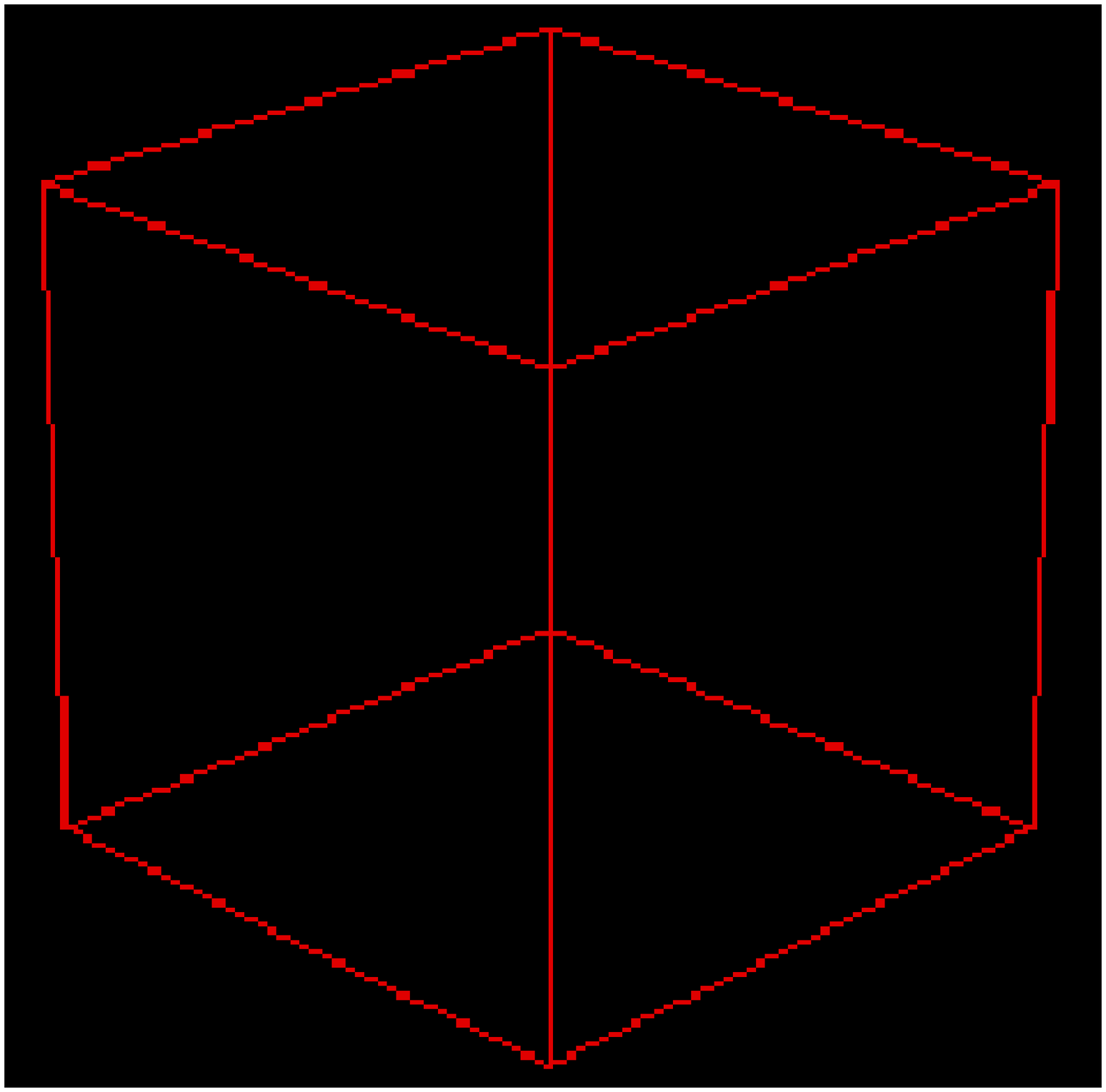}
\caption{The two first images are the geodesic dilatation with a boundary region localized on the ambiguous points such as the SRIO is different. The third image represents the boundary difference of the two previous images. It is empty image since the region localization at the end the growing process is invariant about the SRIO.}
\label{ambi}
\end{center} 
\end{figure}

We suppose in this paragraph that the seeded initialisation follows this order $0,1,\ldots,n$.\\  
To get a boundary localized on the ambiguous points using the SRGMPA, a boundary region is added such as its ZI is always empty. For all the regions except the boundary region, their ZI are localized on the outer boundary region excluding all the regions: $Z^t_i=(X_{i}^t \oplus V)\setminus (\bigcup\limits_{j \in \mathbb{N}} X_{j})$. When a couple $(x,i)$ is extracted from the SQ, there is (see figure~ [\ref{sqb},\ref{ambi}] and algorithm~\ref{alg4}):
\begin{enumerate}
\item p.growth(x, boundary region)) if there is more than two ZI in x and if i = min$\_$elements( pop.Z()[x]),
\item p.growth(x, i)) else
\end{enumerate}
 \begin{algorithm}[h!tp]
\caption{Geodesic dilatation with a boundary as ambiguous points}
\label{alg4}
\algsetup{indent=1em}
\begin{algorithmic}[20]
 \REQUIRE $I$, $S$ , $V$ \textit{//The binary image, the seeds, the neighborhood}
\STATE \textit{// initialization}
\STATE System$\_$Queue s$\_$q( $\delta(x,i)=0\mbox{ if }I(x)\neq 0, OUT\mbox{ else}$, FIFO, 1); \textit{//A single FIFO queue such as if $I(x)=0$ then $(x,i)$ is not pushed in the SQ.}
\STATE Population p (s$\_$q); \textit{//create the object Population}
\STATE \fbox{\textbf{Tribe passive($V=\emptyset$);}}
\STATE \textit{//create a boundary region/ZI, $(X^t_b,Z^t_b)$ such as $Z^t_i=\emptyset$}
\STATE \fbox{\textbf{int ref$\_$boundary   = p.growth$\_$tribe(passive);}}
\STATE \fbox{\textbf{Restricted $N$=$\mathbb{N}$; }}
\STATE Tribe active(V, N);
\FORALL{$\forall s_i\in S$ in the order $0,1\ldots$} 
\STATE int ref$\_$tr   = p.growth$\_$tribe(actif); \textit{//create a region/ZI, $(X^t_i,Z^t_i)$ such as \fbox{$Z^t_i=(X_{i}^t \oplus V)\setminus (\bigcup\limits_{j \in \mathbb{N}} X_{j})$}}
\STATE  p.growth($s_i$, ref$\_$tr ); 
\ENDFOR
\STATE \textit{//the growing process}
\STATE  s$\_$q.select$\_$queue(0); \textit{//Select the single FIFO queue.}
\WHILE{s$\_$q.empty()==false}
\STATE  $(x,i)=s$\_$q$.pop();
\IF{pop.Z()[x].size()$\geq$2 and i= min$\_$elements( pop.Z()[x])}

\STATE \fbox{\textbf{ p.growth(x, ref$\_$boundary); }}\textit{//growth of the boundary region}
\ELSE
\STATE \fbox{\textbf{ p.growth(x, i ); }}\textit{//simple growth}
\ENDIF
\ENDWHILE
\RETURN p.X();
\end{algorithmic}
 \end{algorithm}
This partition is invariant about the SRIO but is not a $V$-boundary-partition (see figure~\ref{sqb}) since there are some holes on the boundary region.

\section{Conclusion}
In discrete space, the boundary definition is not oclearly defined. Using the SRGPA, we have proposed two growing processes to do a simple or V-boundary partition. These growing processes have incertitude on the regions boundary localisation. To overcome this problem, we have defined a set of ambiguous points such as in a discrete space, it is impossible to know to which regions they belong. Knowing that, we have defined a growing process with a boundary region localized on these ambiguous points. The associated partition to this growing process is invariant about the SRIO but it is only a simple since there are some holes on the boundary region. Depending on the algorithm or the application, it is possible to apply a post-treatment to label these ambiguous points to the regions. For example in the case of the evolution of the cement paste microstructure, the ambiguous pixels have been always affected to the void phase. There is an over-localization of this phase but the error due to the over-localization is always the same and can be estimated.
\appendices
\section{Summary of the previous article}
\label{ap:sum}
The idea of the first article is to define three objects: Zone of Influence (ZI), System of Queues (SQ) and Population. The algorithm implementation using SRGPA is focused on the utilisation of these three objects. An object ZI is associated to each region and localizes a zone on the outer boundary of its region. For example, a ZI can be the outer boundary region excluding all other regions. An algorithm using SRGPA is not global (no treatment for a block of pixels) but local (the iteration is applied pixel by pixel belonging to the ZI). To manage the pixel by pixel organisation, a SQ sorts out all pixels belonging to ZI depending on the metric and the entering time. It gives the possibility to select a pixel following a value of the metric and a condition of the entering time. The object population links all regions/ZI and permits the (de)growth of regions. A pseudo-library, named Population, implements these three objects. An algorithm can be implemented easier and faster with this library, fitted for SRGPA.
\section*{Acknowledgment}
I would like to thank my Ph.d supervisor, P. Levitz, for his support and his trust. The author is indebted to P. Calka for valuable discussion and C. Wiejak for critical reading of the manuscript.  I express my gratitude to the Association Technique de l'Industrie des Liants Hydrauliques (ATILH) and the French ANR project "mipomodim" No. ANR-05-BLAN-0017 for their financial support.

\bibliographystyle{plain}
\bibliography {../bibliogenerale}

\begin{thebibliography}{1}

\bibitem{Beucher2004}
S.~Beucher.
\newblock Algorithmes sans biais de ligne de partage des eaux.
\newblock {\em Note interne CMM}, 2004.

\bibitem{Mehnert1997}
A.~Mehnert and P.~Jackway.
\newblock An improved seeded region growing algorithm.
\newblock {\em Pattern Recognition Letters}, 18(10):1065--1071, October 1997.

\bibitem{Najman1996}
L.~Najman and M.~Schmitt.
\newblock Geodesic saliency of watershed contours and hierarchical
  segmentation.
\newblock {\em Ieee Transactions On Pattern Analysis And Machine Intelligence},
  18(12):1163--1173, December 1996.

\bibitem{Schmitt1989}
M.~Schmitt.
\newblock Geodesic arcs in non-euclidean metrics: Application to the
  propagation function.
\newblock {\em Revue \&Intelligence Artificielle}, 3, no.2:43--76, 1989.

\bibitem{Tariel2008b}
V.~Tariel.
\newblock Conceptualization of seeded region growing by pixels aggregation.
  part 1: the framework.
\newblock {\em submitted}, 2008.

\end{thebibliography}
\end{document}